# One Size Does Not Fit All: Quantifying and Exposing the Accuracy-Latency Trade-off in Machine Learning Cloud Service APIs via Tolerance Tiers

Matthew Halpern[*] Behzad Boroujerdian[*†] Todd Mummert[‡] Evelyn Duesterwald[‡] Vijay Janapa Reddi[*†]

[*]The University of Texas at Austin    [†]Harvard University    [‡]IBM Research

*Abstract*—Today's cloud service architectures follow a "one size fits all" deployment strategy where the same service version instantiation is provided to the end users. However, consumers are broad and different applications have different accuracy and responsiveness requirements, which as we demonstrate renders the "one size fits all" approach inefficient in practice. We use a production grade speech recognition engine, which serves several thousands of users, and an open source computer vision based system, to explain our point. To overcome the limitations of the "one size fits all" approach, we recommend *Tolerance Tiers* where each MLaaS tier exposes an accuracy/responsiveness characteristic, and consumers can programmatically select a tier. We evaluate our proposal on the CPU-based automatic speech recognition (ASR) engine and cutting-edge neural networks for image classification deployed on both CPUs and GPUs. The results show that our proposed approach provides a MLaaS cloud service architecture that can be tuned by the end API user or consumer to outperform the conventional "one size fits all" approach.

## I. INTRODUCTION

*Machine Learning as a Service (MLaaS)* platforms allow developers to incorporate "intelligent" functionalities, such as image recognition, speech transcription, and natural language comprehension, into their applications. When an application needs to invoke one of these functionalities they make a request using the MLaaS API – offloading the computation to the cloud. This paradigm provides an attractive model for cloud service providers, motivating prominent technology firms that include Amazon, Google, IBM, and Microsoft to deploy and operate their own platforms.

Today's MLaaS cloud service architectures follow a "one size fits all" deployment strategy. Multiple instantiations of the same service version are scaled out across the service's computing infrastructure to handle all of its users. This design is problematic because MLaaS relies on computations that are of a statistical nature: deeper exploration yields more accurate results but also requires more processing time to perform. As a result, *MLaaS providers are forced to make an explicit trade-off between the service's result accuracy and responsiveness*.

However, API consumers are broad and different applications have different *accuracy, responsiveness, and monetary budget* constraints. On one hand, there are accuracy-critical application domains, such as those in healthcare and finance. Inaccurate results can have large financial, and even fatal, consequences in these domains. Increases in service response time can be justified if it can provide the best result possible. On the other hand, there are responsiveness-critical application domains, such as social networking and shopping where slow user experiences in these domains lead to poor user engagement, so the results only matter if they can be procured before the user abandons the applications. Some result is better than no result. Additionally, some applications are also cost-critical, or pricing sensitive. Providing these capabilities is not sustainable to the developers if it is not cheap enough to profit. API consumers pay per use of the cloud service API each time it is invoked – cutting into their application's revenue.

So, the industry sees a strong *disconnect between the individual needs of API consumers and the way machine learning-based cloud services are being deployed.* Though API consumers possess diverse operational requirements, the "one size fits all" deployment strategy has no means to individually adapt to each of them. The MLaaS providers are forced to make a static design-time decision based on generic needs. Hence, there is a need to rethink the way we design, deploy, and operate MLaaS services, taking both MLaaS users and service providers into account in a mutually beneficial way.

In this paper, we quantitatively demonstrate the limitations of the one size fits all model, using automatic speech recognition and image classification MLaaS deployments as two representative examples. We study different service versions to show that the limitations of the one size fits all model is independent of the hyperparameters that are used to tune the model. We further demonstrate that the inherent latency-accuracy trade-off in MLaaS implies that to improve the accuracy needs of some "extreme" queries, the overall or average latency in a single monolithic model has to grow significantly.

Therefore, to address the limitations of the one size fits all approach, we introduce *Tolerance Tiers* for MLaaS platforms. Similar in spirit to virtualized computing platforms, such as Amazon EC2, that allow customers to select computing resources optimized for different application criteria (e.g. performance, storage), our goal with Tolerance Tiers is to enable API consumers to programmatically configure the MLaaS to act in accordance with their operational requirements, which might be accuracy, responsiveness, cost or something else.

Tolerance Tiers ensembles multiple versions of a machine learning-based service to compute a result. The rationale is that the *multiple model versions can be combined together in ways that provide better accuracy, responsiveness, and cost trade-offs than can be achieved by any single service version on its own*. Tolerance Tiers are able to provide a more fine-grained trade-off space than if only using a single service version at a time to produce a result. Leveraging pools of the different service versions, the MLaaS load balancer employs intelligent routing policies that dictate how, and when, a service version will be used to process a given service request depending on the specific Tolerance Tier.

Also, Tolerance Tiers enable consumers to sacrifice the service's result quality to improve other aspects of the service, such as the service response time and invocation cost. Each tier has two parts: an optimization objective and an error tolerance. The tolerance provides a guarantee as to how the particular Tolerance Tier performs relative to the most accurate known Tolerance Tier in terms of result accuracy while improving the service in accordance with the optimization objective. For example, given a sufficient number of requests to the 1% Tolerance Tier its error is statistically guaranteed to be less than 1% worse of the most accurate Tolerance Tier the service can provide if those requests had been made to it. At the same time, the service should provide noticeably better response times and/or cheaper invocation costs as compared to the most accurate tier.

To alleviate the service provider from manually creating the routing policies for each Tolerance Tier, we also provide a training framework. The framework constructs routing policies that aggressively optimize each Tolerance Tier while maintaining its corresponding statistical guarantees. Using a CPU-based production-grade automatic speech recognition (ASR) engine that has been in service and cutting-edge neural networks for image classification deployed on both CPUs and GPUs, we show that *Tolerance Tiers* can achieve (1) design generality; (2) accuracy guarantees; (3) response time improvements; and (4) cost improvements. Tolerance Tiers is designed to leverage general characteristics of machine learning models (i.e. the latency-accuracy design trade-off and result confidence metrics). While the ASR and image classifications have fundamentally different designs in terms of machine learning, the same trends can be observed for both of them. We observe no accuracy degradation violations throughout the evaluation of Tolerance Tiers, demonstrating that guarantees provided by our automatic routing rule generation framework are upheld. Tolerance Tiers enables service latency reductions of 19% for a 1% accuracy tolerance, 45% for a 5% accuracy tolerance, and 60% for a 10% accuracy tolerance. Last but not least, Tolerance Tiers enables invocation cost reductions of 21% for a 1% accuracy tolerance, 60% for a 5% accuracy tolerance, and 70% for a 10% accuracy tolerance. In short, our effort includes:

- **Limitations of "One Size Fits All"** We quantitatively show that the conventional cloud deployment solution conflicts with the diverse needs of machine learning-based cloud service API consumers.
- **Tolerance Tiers** We propose an alternative cloud service architecture that allows API consumers to specify their accuracy-latency requirements as opposed to the cloud service provider at design time–shifting the decision to the party that is more informed to make it and also impacted by it.
- **Service Version Ensembling** We combine various service version ensembling schemes that enable better accuracy, responsiveness, and cost trade-offs than selecting a single version to compute a result.

We introduce our applications in Sec. II. Using these applications, we quantify the limitations of the "one size fits all" approach in Sec. III. Sec. IV introduces *Tolerance Tiers*, which we evaluate in Sec. V. We present prior work in Sec. VI and conclude in Sec. VII.

## II. APPLICATION DOMAINS

We study speech and vision applications. We describe automatic speech recognition and image classification. We select them because of their widespread use among the machine learning-based cloud services in use today.

### A. Automatic Speech Recognition (ASR)

Automatic Speech Recognition (ASR) converts human speech into human-readable text. Given an *utterance*, or an human speech sample, an ASR engine seeks to identify what words were spoken. In its simplest form, ASR is a graph-based search problem. The ASR engine breaks the speech down into regularly segmented intervals of speech, known as frames. The engine then calculates metrics, or features, about each frame which are then fed into a neural network that generates a model for different aspects of the audio such as speaker pronunciation and environmental conditions of the recording. This acoustic model is combined with a language model, which encapsulates the word semantics and other grammatical aspects of the spoken language, are combined to form a hidden markov model (HMM), creating a graph-based, probabilistic representation of human speech.

Searching the entire HMM is expensive to perform in its entirety given the complexity of human speech. Instead, an approximate heuristic-driven beam search is used. Conceptually speaking, the beam search heuristics dictate the subset of the HMM searched controlling the breadth and depth of the search. Therefore, the *heuristic-driven nature of beam search imposes a critical accuracy-latency trade-off in ASR engine design*. This trade-off is well-established in modern ASR engines [1], [2], [3], [4], [5], [6]. The search's accuracy is directly proportional to the search space size whereas the latency is indirectly proportional to it. Further, the beam search itself is only guaranteed to produce a locally optimal result because only a subset of the HMM is searched.

We use a production-grade ASR engine which follows the design of state-of-the-art ASR engines from Baidu [1], IBM [2], Microsoft [3], and Google [7]. The engine uses a *heuristic-driven beam search approach, which imposes a critical accuracy-latency trade-off in ASR engine design*. Searching the entire HMM is expensive to perform in its entirety given the complexity of human speech. Instead, an approximate heuristic-driven beam search is used. This trade-off is established in modern ASR engines [1], [2], [3], [4], [5], [6].

To evaluate ASR, we use transcribed utterances from the VoxForge open-source speech transcript repository [8] to benchmark the ASR service. The dataset consists of over 35,000 utterances that together make up 53 hours of audio time and feature over 3,500 speakers and across different recording environments. We use word error rate (WER) to evaluate accuracy. WER is a well-established metric to assess ASR transcription accuracy, where a lower WER indicates a more accurate transcription. The WER for an utterance $u$ is the ratio of word errors (i.e. insertions, deletions, and substitutions) between the ASR engine's hypothesis, $Hyp(u)$, and its reference transcript, $Ref(u)$ to the number of words in the reference transcript:

$$WER(u) = \frac{\mid WordErrors(Hyp(u), Ref(u)) \mid}{\mid Ref(u) \mid}$$

### B. Image Classification (IC)

Image Classification (IC) is done using convolutional neural networks (CNN) [9]. The input to the CNN propagates through *layers* of *neurons* which computationally correspond to a series of matrix multiplications. The coefficients for each matrix are generated through a training process. Nonetheless, *the number of layers and neurons within each layer imposes an accuracy-latency trade-off in neural network design*. Scaling up the neural network increases accuracy, but increases the amount of processing necessary to compute.

We use 45,000 images from the ILSVRC2012 validation set [10], which spans 1,000 different image categories. In both cases, these datasets were not used for training, thereby eliminating any training bias in our prediction results. To evaluate accuracy, we use the *top-1 error*. The top-1 error corresponds to whether the class with the highest probability in the output layer (i.e. the argmax) is the image's actual label. Unlike the WER whose value falls in a continuous range (i.e. 0% to 100% typically), the top-1 error is a binary condition. The top-1 error is either 0% or 100% depending on whether or not the argmax class and actual class are the same.

## III. THE "ONE SIZE FITS ALL" LIMITATION

Conventional cloud service architectures consist of a single machine learning model used to process all service requests. Under this "scale-out" design, all service requests from API consumers are processed by the same model, instantiated across different service nodes. We examine the shortcomings of a "one size fits all" approach using our ASR and IC services. We study different versions of the service that encompass the pareto-optimal accuracy-latency trade-off space, where we test the services with representative user requests. While both services exhibit the latency-accuracy design trade-off, they manifest in different ways. For the ASR service, we consider the accuracy-latency trade-off for the heuristic driven beam search (as explained in the previous section) and for the image classification service we study different neural network architecture implementations. We demonstrate that given the accuracy-latency trade off presence across various models, inflexibility of any approach that resorts only to one model renders it inefficient across applications with diverse needs.

### A. Inherent Trade-off Due to Model Versions

Models have accuracy-latency trade-offs. We explore the trade-offs by considering different versions.

**ASR Model Versions:** We consider seven versions. Each version uses a different set of heuristic parameters. Conceptually speaking, these heuristics are the product of two orthogonal concerns. The first is the hypothesis pruning policies, which discard all but the top $N$ most probable hypotheses and restrict the search space size. The second is the scope pruned: a single hypothesis (i.e. local), a branch of hypotheses (i.e. global), and the entire HMM (i.e. network). These combinations lie along the ASR engines accuracy-latency Pareto frontier, which was produced from exhaustively sweeping (i.e. grid search) of the heuristic values. Six beam search heuristic parameters were swept by the ASR engine experts that optimize the system.

**IC Model Versions:** We evaluate cutting-edge neural networks used for image classification. Specifically,

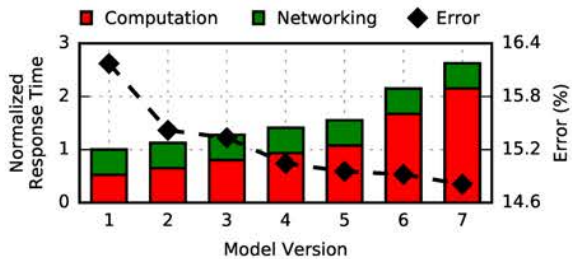

(a) Automatic Speech Recognition (ASR) Service Trade-off.

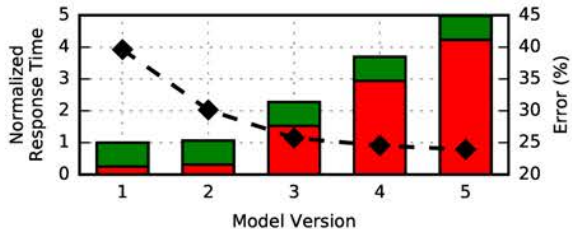

(b) Image Classification (IC) Service Trade-off.

Figure 1: Error and response times across the different cloud service versions for ASR and IC. The different model versions expose a notable accuracy-performance trade-off. In both cases, the service's accuracy can be dramatically improved at the expense of client response time. Server-side processing can dominate the service's end-to-end response time in comparison to networking.

we use models posted on the Caffe Model Zoo [11] for the 2012 ImageNet Large Scale Visual Recognition Competition (ILSVRC2012) [10]: SqueezeNet [12], GoogleNet [13], and the 50-, 101-, and 152-layer ResNet [14] Networks. We also evaluated AlexNet [15], the 16- and 19-layer VGG [16] networks but do not study them as they do not lie within the pareto-optimal accuracy-latency frontier corresponding to the other models. While there are many nuanced differences between these different networks, the key differences we focus on is the number of parameters within each model. For example, the 152-layer ResNet network has an order of magnitude more layers than SqueezeNet and several magnitudes more neurons. As a result, the ResNet is more accurate than SqueezeNet, but is significantly more computationally intensive.

### B. Accuracy vs. Latency Trade-off in Models

Fig. 1 shows the accuracy-latency trade-offs across ASR (Fig. 1a) and IC (Fig. 1b). As mentioned previously, we use 36000 transcribed utterances from the VoxForge open-source speech transcript repository [8] and 45000 labeled images from the ILSVR2012 validation set [10]. In both cases, these datasets are sufficiently large and diverse to use for our analysis. Additionally they were not used to train either service, thereby eliminating any sources of prediction biases.

**Impact on Accuracy** The lines in Fig. 1a and Fig. 1b shows the error reduction across the service versions for ASR and image classification, respectively. Each point corresponds to the mean error for each configuration, which has been sorted from greatest to least. We use this enumeration to refer to each configuration throughout the paper. For the ASR service in Fig. 1a, the beam search configuration used can reduce the error by 9.2% (from 16.17% to 14.8%). The image classification in Fig. 1b is more sensitive to the version used as the most accurate version yields a 65.6% error reduction over the least accurate version (39.6% to 23.9%).

**Impact on Client Latency** Fig. 1 shows the service's end-to-end response time breakdown. We instrument the cloud services quantify the extent to which the service version used increases the service response time. Specifically, we isolate the server-side processing time (red) from the network transmission time (green) throughout the total response time.

Response time increases across the service versions. Fig. 1a shows the response time increases by 2.6×. While the networking time is non-negligible, server-side processing quickly becomes the response time bottleneck. The ASR engine consumes over 50% of the service response time at configuration one and steadily increases to over 80% by configuration seven. Similar trends can be observed for the image classification service in Fig. 1b where the response time increases by 5× and server-side processing has a comparable level of prominence.

### C. Accuracy vs. Latency Sensitivity to Input Categories

Given the accuracy-latency tradeoff presence in the models, a naive approach may want to map each model to a *Tolerance Tier*. However, such an approach will not provide the accuracy guarantees needed for each of the tiers because *models are input sensitive, i.e. their accuracy varies based on their input*.

To demonstrate input sensitivity, we analyze the service requests to understand how they are each individually affected by the different service versions. We show how inputs can be categorized on their accuracy-latency behavior into four behavior classifications. Using measured requests to the ASR service as examples, we describe each category and their implications to service provider design. Given such input diversity, we conclude that a "one size fits all" approach has to compromise and make accommodation for all inputs in one model.

**Result Quality Improves:** This category follows the intuition that increased processing time yields accuracy improvements. Fig. 2a shows an example ASR service request from this category. The error is plotted on the *y*-axis and the service response time, normalized to configuration one, is shown in the *x*-axis. Configuration one produces an error of 24.6% that decreases to 23 at configuration two and finally to zero at configuration six. The accuracy improvement does come at the expense of more than doubling the service response time. Yet, it is justifiable because the service's result quality improved.

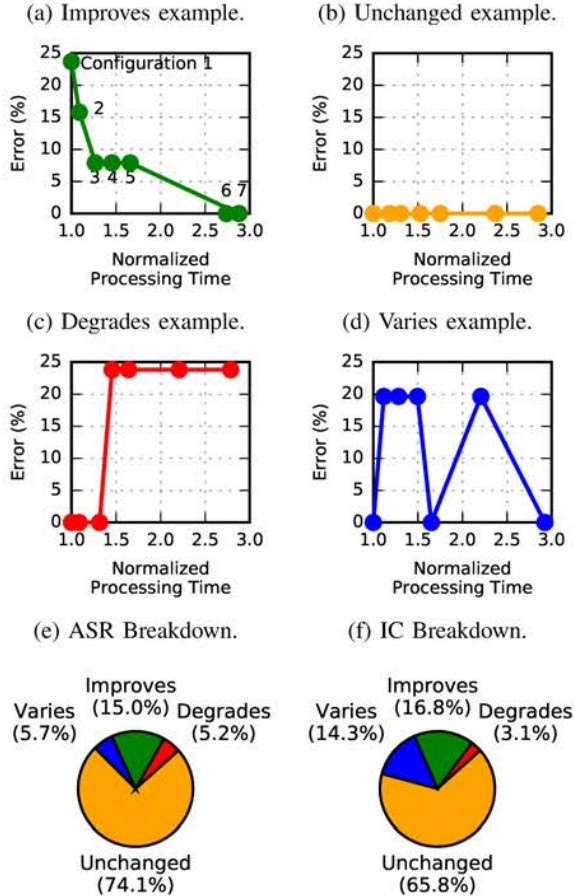

Figure 2: Accuracy-latency behaviors across different service versions. We identify four distinct behavioral categories. Intuitively speaking, the result accuracy should either improve with increased processing time (a) or stay the same (b). However, we observe this is not always the case as the increased processing time can induce more errors for some requests (c) or exhibit unpredictable patterns (d). Therefore, the service version needed for each request can vary, and even conflict for both the ASR (e) and image classification (f) services.

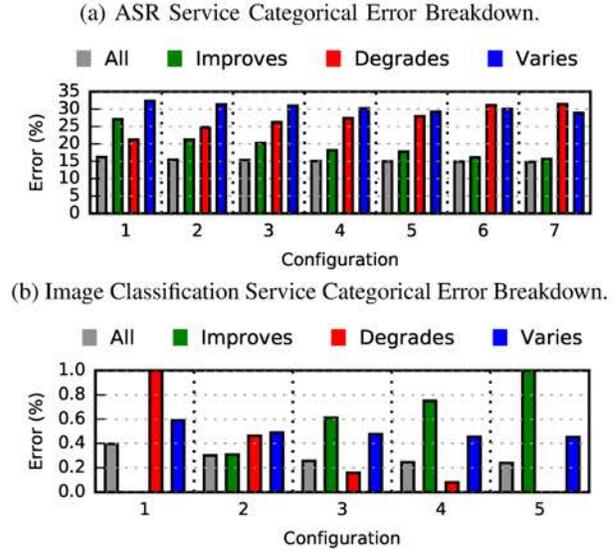

Figure 3: Error for the different behavioral categories. While the service's error holistically improves across the different service versions, many request results worsen.

*Result Quality Unchanged:* There are cases where increasing processing time does not change result accuracy. Fig. 2b shows an utterance whose error remains at zero throughout all of the ASR service versions. This category represents the requests whose result accuracy is unaffected by *any* the service version used. Increased processing time only leads to excessive computation, as the result quality remains unchanged.

**Result Quality Degrades:** Counterintuitively, increased processing time can degrade quality. Fig. 2c shows an ASR service request where configuration one produces the most accurate transcription. Subsequent service versions introduce error while also increasing the response time. This phenomenon occurs because the beam search does not guarantee (globally) optimality when it computes a result. A large search space means the beam search is more likely to produce an erroneous transcription as the heuristics make the beam search too liberal. For IC, a similar phenomenon is seen for neural networks as each network experiences some degree of overfitting to its training set and different networks will overfit in different ways. This category is least desirable because any increase in processing time is not only excessive, but also degrades quality.

**Result Quality Varies:** Some requests do not exhibit clear accuracy-latency characteristics, as shown in Fig. 2d. The relationship between processing time and accuracy is non-monotonic. Configurations one, five, and seven are all able to provide the most accurate result for ASR service request, but the best configurations are arbitrary for service requests in this category. This phenomenon also occurs because of the non-optimality characteristics of machine learning techniques. There is no clear design approach for these utterances because they do not bias a particular configuration.

**Result Quality Summary:** Fig. 2e and Fig. 2f show the breakdown of requests subject to the accuracy-latency behavior categories for the ASR and image classification services, respectively. Almost an outstanding majority of the service requests are not sensitive to the service version used. Over 74% and 65% of the service requests fall into this category for the ASR and image classification services respectively. For both services, over 15% of the requests belong to the improves category. The image classification service has a more notable number of variable requests. The summarizing takeaway of Fig. 2 is that no one service version provides the best result quality for all service requests to either of the study. The "one size fits all" approach chooses to benefit certain accuracy-latency categories over others.

### D. Comprehensive Accuracy vs. Latency Analysis

We extend our analysis to include all of the service requests we have studied thus far. Specifically, we quantify the contention between these different accuracy-latency categories under the "one size fits all" approach.

Fig. 3 shows the error for the different categories across the service versions for ASR (Fig. 3a) and IC (Fig. 3b). The "unchanged" group is not shown because it is unaffected by the configurations. We show the error for all service requests in the bars labeled "all," which allows us to see how the different heuristic configurations impact the service's result quality.

The "all" bar shows that accuracy improves across the configurations for both services (i.e., considering all service requests), which happens for two reasons. First, as shown in Fig. 2e and Fig. 2f, the majority of utterances belong to the "unchanged" category, so a small number of service requests see variances in result quality. Second, amongst the requests whose result qualities vary, most belong to the "improves" category. To a lesser extent, the "varies" category also benefits from the longest running configuration.

### E. Limitations of the "One Size Fits All" Summary

We presented three key findings in this section. First, our results show that a large accuracy-latency trade-off space exists for both the speech and vision application services. A 2.6× increase in response time can reduce the ASR service's error by over 9% and a 5× response time increase reduces the image classification service's error by over 65%. Second, our analysis shows that *due to the differences among the inputs, no one service version is best-suited to process all service requests, and as such a "one size fits all" model that seeks to satisfy a quality constraint must surrender to the small portion of inputs that demand better models for improvement, and hence it must compromise the response times of all the inputs*. Third, our analysis implies that if API consumers wish to have flexibility in accuracy-latency trade-offs and optimize for some objective function (e.g., cost or responsiveness), *service providers should adopt policies with which ensemble of models with various invocation schemes (called policies) are selected dynamically to satisfy the tiers' accuracy need*.

### IV. TOLERANCE TIERS' ARCHITECTURE

We propose *Tolerance Tiers*, an alternative model for machine learning-based cloud services, where a set of tiers with different accuracy/latency characteristic are provided to the consumers to select from. Using Tolerance Tiers, consumers are empowered to make trade-offs between the cloud service's accuracy, response time, and cost as compared to the rigid, conventional cloud service architectures that consist of a single machine learning model used to process all incoming service requests.

Similar to how Amazon Web Services (AWS), Google Cloud Platform Services, Microsoft Azure Cloud etc. allow customers to select different machine instance types for a price, Tolerance Tiers allows API consumers to programmatically sacrifice result quality to improve other aspects about the service, such as the service response time and invocation cost. When making a request to the service, the API consumer selects a specific Tolerance Tier to process the request. The Tolerance Tier specifies a lower bound on the service's expected accuracy relative to the best accuracy that can be achieved. The more aggressive the Tolerance Tier, meaning the higher degree of error that can be tolerated, the larger the opportunity for improvement in other aspects of the service.

### A. Service Request Annotation

A Tolerance Tier service request is shown below, which resembles how most vendors make their services accessible to users. Assume that the endpoint performs ML computation and returns the result. In addition to the input file for the service to process, the API consumer annotates the request with two additional headers.

```
curl --header Tolerance: 0.01
     --header Objective: response-time
     --data-binary @input-file-name
     -X POST http://cloud-service/compute
```

First, the API consumer must specify an acceptable result Tolerance. This corresponds to the relative result quality degradation as compared to the most accurate version the service can provide. For example, given a sufficient number of requests to the 1% tolerance tier its error is statistically guaranteed to be less than 1% worse of the most accurate tolerance tier the service can provide if those requests had been made to it.

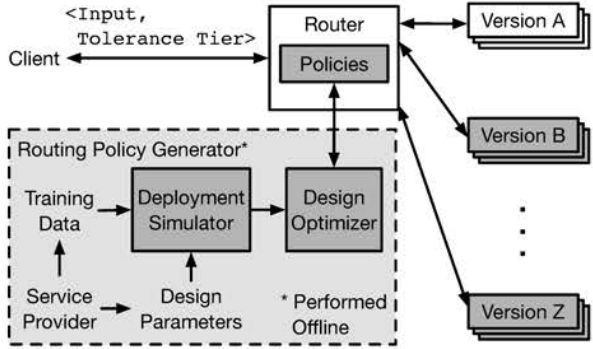

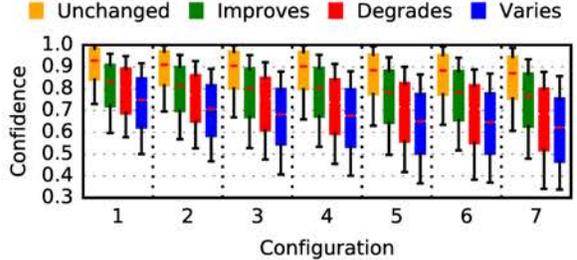

Figure 4: *Tolerance Tiers* Design and Implementation Overview. The end-to-end cloud service architecture, which required simple modifications to the original service architecture (in gray), deploys multiple pools of different service node versions together managed by the service's load balancer. The API consumer annotates service requests with the *Tolerance Tier* best-suited to their operational objectives. The service's load balancer is programmed to enact intelligent routing policies based on the specified *Tolerance Tier*. A routing policy generator alleviates the service provider of having to devise the appropriate routing policies manually, while providing high statistical guarantees of its production performance.

Figure 5: Distribution of the transcription confidences for the different behavioral categories subject to the different ASR service versions. Transcription confidence can be used to distinguish between different categories.

Second, the API consumer must specify an optimization *Objective*. In the example above, the "response-time" *Objective* minimizes response time while adhering to the specified accuracy Tolerance. In addition to the "response-time" objective, we also consider a "cost" objective (i.e., Objective: cost) that seeks to minimize the service's invocation cost by minimizing the service's server-side processing time.

*B. Service Version Ensembling*

Each Tolerance Tier employs a service version ensemble to process a service request, as shown in Fig. 4. We deploy two complementary service node versions: one tuned for accuracy and another for response time. This allows us to deploy service node versions that can specifically target the behavioral categories identified in Sec. III-C. The accuracy-oriented version provides the best accuracy for the utterances from the "improves" category (and many from the "varies" category), while the speed-oriented version provides the best accuracy for the "degrades" category. Additionally, the "unchanged" category observes a speed benefit from the speed-oriented version as well, but can be run on either version without any loss in accuracy.

Based on our experimental analysis using the previously studied data shows that the benefits of ensembling multiple service versions greatly diminishes with more than two versions. So, from hereon forward, in the paper we only consider deploying at most two service versions.

To utilize multiple versions at once, we must determine which version is the most likely to provide the most accurate result. We observe that many machine learning algorithms provide metrics that can serve as suitable heuristics to estimate result quality. Many of the cloud-based machine learning offerings provide confidence scores with the result they return [17], [18], [19].

Fig. 5 shows the plots corresponding to the confidence distributions for each of the categories subject to the beam search heuristic configurations. There are two points we would like to emphasize. The first is that the "unchanged" utterances exhibit significantly higher confidences than the other categories throughout all of the configurations, which demonstrates that many of these utterances can be easily identified. The second is that the "degrades" category have noticeably lower confidence scores than the "improves" case. Therefore, once the "unchanged" utterances have been filtered away, we can easily distinguish between these two categories to some confident degree.

It is worth noting that confidence, and other similar metrics that imply result quality, are common computed within many machine learning algorithms. For example, the output layer of a neural network consists of a score for each possible class (the class with the highest score is typically the one deemed as the predicted result) and the k-nearest neighbors algorithm uses a distance metric to reach a consensus on what result to return. Hence, the approach we are proposing is generalizable.

*C. Routing Policies*

Our service cluster uses routing policies to handle incoming client requests. In addition to the original "one size fits all" approach, we implement two heuristic-based policies for our router. These policies require a few lines of code to implement within the service's load balancer. The insight behind the proposed policies is that we can make trade-offs not only between the service's result quality and response time, but also its *invocation costs*. Cloud service providers charge API consumers for machine processing time. For example, IBM charges $0.02 cents a minute for their ASR processing time at the time of writing [20]. Therefore, a result that takes longer to compute results in *higher invocation costs* as the computing resources must be used for a longer amount of time to process the result. Depending on the service versions and how the following policies utilize them, the invocation cost can vary dramatically. Next, we describe the individual routing policies. We use configuration one and seven for the ASR service, as they offer complementary accuracy-latency trade-offs.

**One-Size-Fits-All (OSFA):** This is the deployment where a single node version is used to handle all client requests. We include this for comparison. We see how each policy operates in Fig. 6. Fig. 6b shows the timeline for the OSFA policy, where each bar corresponds to the time spent in each of the cloud components described in Sec. V-A. The OSFA policy forwards the request to any service node upon receiving the request. The node returns a result to the router and it responds back to the client. We use configuration seven because it is the most holistically accurate, but slowest configuration.

*Sequential Issue (Seq):* Fig. 6c shows the timeline for the Seq policy. The router first dispatches the request to one of the service node versions. Once the node returns its result to the router, the router has a decision to make. If the confidence is above a preset threshold, the policy can decide to return the result to the client. Otherwise, it is sent to another service node version for additional processing. We distinguish between these cases as Early Termination (ET) and Full Operation (FO), respectively. While Fig. 6c shows that we dispatch to configuration seven before configuration one, there are situations where dispatching to configuration one before configuration seven is ideal.

*Concurrent Issue (Conc):* Fig. 6d shows how Conc duplicates the request and forwards it to multiple service node versions to process simultaneously. When the faster version returns its result, the router decides to initiate ET if the result confidence is high. If so, the router sends the result back to the client and then terminates the other service node. Otherwise, it waits for the other node to finish processing. Once its result is returned it decides which one to send back by looking at its confidence.

*Trade-off between OSFA, Seq, Conc (and others):*

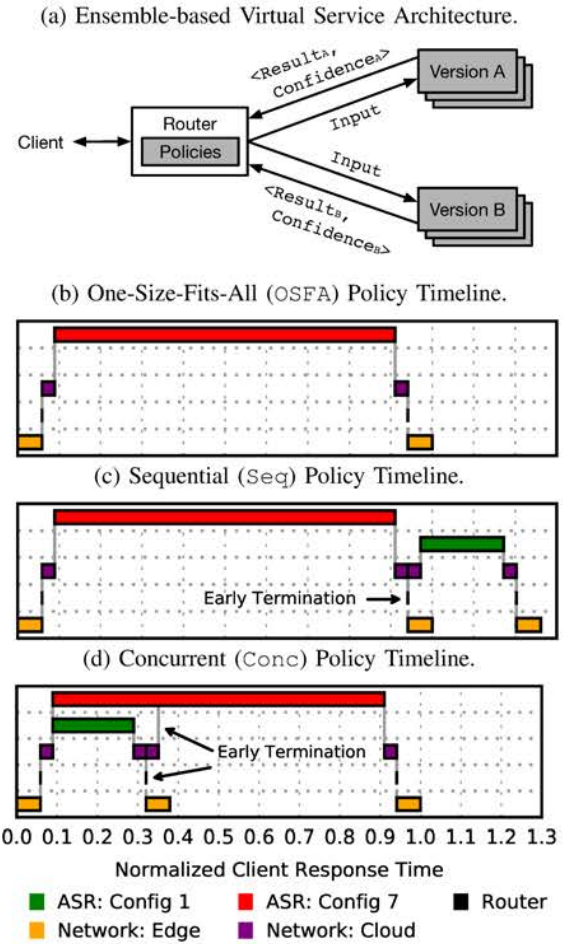

Figure 6: Service version ensemble architecture. (a) Each *tolerance tier* consists of combining multiple service versions together to process a given API request. The load balancer employs intelligent routing policies (b-d) that can dynamically determine which service version within the ensemble is best-suited to produce a result by evaluating result confidences while optimizing for service response time and cost.

The Seq and Conc policies improve accuracy at the expense of response time and IaaS costs as compared to the OSFA approach. In the best case (i.e. ET), Seq can perform as well as OSFA. The critical path for ET in Fig. 6c shows the same behavior as the OSFA in Fig. 6b. However, the response time and cost increase by 20% and 25% respectively when FO is performed. However, this can be tolerable if Seq performs FO as little as possible.

In comparison to OSFA and Seq, the Conc policy sacrifices IaaS costs, using two versions and more machines, to not degrade the service's response time. In the worst case (i.e. FO), the response time will be the same as OSFA with configuration seven. However, the response time can improve by more than 60% when `ET` is used. Similarly, the costs for `ET` are 50% less that `OSFA`, but the portion devoted to configuration seven, which never finished processing, is excessive. This is because configuration one and seven concurrently execute. In `FO`, the IaaS cost for `Conc` is the same as `Seq` because both service node versions will compute the results in either case.

We evaluated more complex solutions including using more than two versions and also a ML-based router; however the simple policies that we discuss here outperformed them. So, we do not include their discussion.

### D. Routing-Rule Generator

Identifying the correct parameters for a Tolerance Tier ensemble can be challenging and cumbersome for the service provider to perform. The design space for service version ensemble is large because it can be deployed with different node versions and routing policies.

Hence, we automatically generate routing rules for the Tolerance Tiers. Specifically, our framework identifies the best-suited parameters to deploy for each Tolerance Tier constraints. Fig. 4 illustrates the framework. The service provider only needs to input training data. Weassume that the training data is representative of future client request traffic. Our assumption is that the service provider has put in the time to carefully produce datasets of their users that are representative of what will be observed in production. Major IT firms already do this. As a best effort to create diversity, we put forth our best effort to consider potential variations in service requests by using 10-fold cross validation in our evaluation, which is standard practice in these situations.

The rule generator uses statistical techniques to construct routing policies with confidence. Specifically, the generator uses bootstrapping [21] to simulate the benefits of each service ensemble configuration under a variety of different scenarios. This approach allows the generator to gain high confidence in the worst-case performance of the service. The code is shown in Fig. 7. The intuition is to simulate each Tolerance Tier configuration with a random subset of the training data to elicit enough performance variation to establish confidence on the worst-case performance it will have in production.

The provider imports and instantiates the `RoutingRuleGenerator` class with training data, a set of candidate service version ensemble parameters to consider, and a confidence score. The rule generator then bootstraps each configuration using the training data (i.e. the `bootstrap` function). The generator then conducts a *trial* where a subset of training data is sampled and simulated. The

```python
from numpy import argmin
from random import choice
from scipy.stats import ppf, zscore
from toltiers.simulator import simulate

class RoutingRuleGenerator:

  def __init__(self, train_data, cfgs, conf):
    self.cfgs = cfgs
    self.conf = conf
    self.results = [
      self.bootstrap(cfg, train_data, conf)
      for cfg in cfgs
    ]

  def bootstrap(self, cfg, train_data, conf):
    trials = []  # tuples of (err_deg, resp_time, cost)
    while any([
        not confident(metric) for metric in zip(*trials)
    ]):
      sample = choice(train_data, k=len(train_data) / 10)
      trials.append(simulate(sample, cfg))
    return [max(metric) for metric in zip(*trials)]

  def confident(self, vals)
    zscores = zscore(vals)
    stdevs = ppf(self.conf)
    return (
      (min(zscores) < -stdevs and max(zscores) > stdevs)
      or (max(zscores) - min(zscores) > 2 * stdevs))

  def generate(self, tols, obj):
    rules = {}
    for tol in tols:
      best_cfg = argmin(map(self.results, obj))
      rules[tol] = self.cfgs[best_cfg]
    return rules
```

Figure 7: Routing Rule Generator code. The generator uses *bootstrapping* to sample and simulate different service version ensemble configurations across different service request load scenarios. This allows the generator to construct service version ensembles with a high degree of statistic confidence.

generator continues sampling and conducting trials until the generator reaches specified confidence with the observed error degradations, response times, and costs from the trial simulations. The worst-case error degradation, response time, and cost are recorded. After all the possible design configurations have been bootstrapped, the service provider can generate routing rules by specifying Tolerance Tier ranges to deploy and a corresponding objective (i.e. the `generate` function).

### V. EVALUATION

We evaluate two policies: to minimize each service's response time and API invocation cost. We consider tolerance degradations up to 10% in 0.1% intervals reinforced with a 99.9% confidence. Our goal is not to pick a single arbitrary sweet-spot to focus upon. Instead, we present the holistic design space that API consumers and service providers can exploit for making different trade- offs. This allows us to show that Tolerance Tiers (1) does not violate its accuracy guarantees while (2) decreasing service response time and (3) reducing invocation costs.

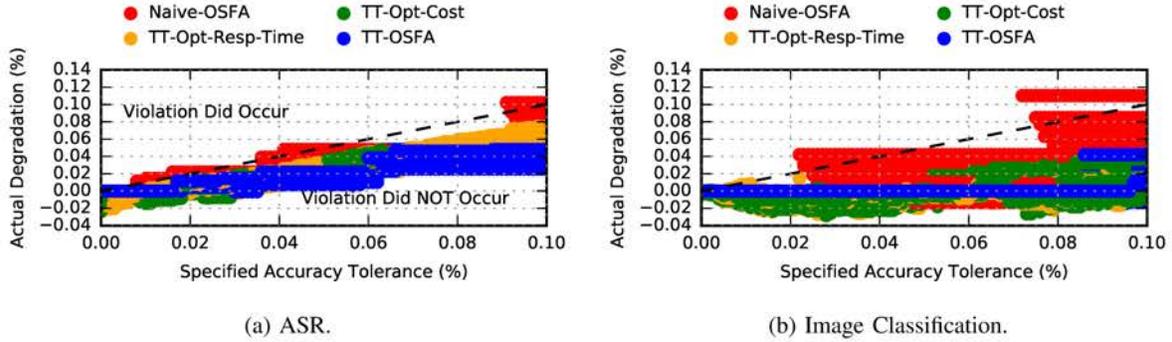

(a) ASR.   (b) Image Classification.

Figure 8: Error results, where each point represents the specified (i.e. guaranteed) tolerance, *x*, versus the actual error degradation that occurred, *y*. The dashed line corresponds to the violation boundary. Dots above the line is a violation as the error degradation is greater than the tolerance. There are no violations with Tolerance Tiers.

### A. Experimental Setup

For the evaluation, we add an additional 11% more service inputs to study each service, bringing the total to 40,000 and 50,000 for the ASR and image classification service, respectively. We then use ten-fold cross validation to evaluate the techniques where each fold consists of randomly selecting 90% of the service requests input to train the rule generator and then the remaining 10% of the service requests for the evaluation. Unless otherwise noted, the results correspond to the mean of all ten folds.

We deploy the ASR service on IBM Cloud [22], a cloud provider comparable to Amazon Web Services [23]. We provision a pool of 64 virtualized Docker container [24] instances. Each instance is provisioned with four Intel Xeon processor cores, 8 GB of memory, and 25 GB of disk space. We use Netflix Zuul [25] for load balancing, which is used by many well-established companies [26]. The image classification service runs bare-metal in a private cloud computing environment. An instance runs an eight-core Intel Xeon processor with 64 GB of memory, 500 GB of storage, and a companion NVIDIA K80 GPU. NGINX is used as the load balancer [27].

### B. Policies

We consider the following policies for our evaluation:

- `Naive-OSFA` is using the *entire* training set to pick the best single service version to deploy for the service subject to the specified accuracy tolerance.
- `TT-OSFA` is using our Tolerance Tiers statistical framework to choose the best single service version to deploy that will provide high statistical guarantees to not incur an accuracy tolerance violation.
- `TT-Opt-Resp-Time` is using Tolerance Tiers to minimize the service response time.
- `TT-Opt-Cost` uses Tolerance Tiers to minimize the service invocation costs (i.e. machine time).

### C. Accuracy Degradation

There are no formally established Service Level Agreement (SLA) definitions for MLaaS. So, we use a relative quality degradation metric to evaluate Tolerance Tiers. We show that accuracy tolerances guaranteed by Tolerance Tiers are respected throughout our analysis.

Fig. 8a and Fig. 8b show expected versus actual result quality degradation for the ASR and image classification services, respectfully. Each point represents the specified (i.e. statistically guaranteed) tolerance, *x*, versus the actual error degradation that occurred, *y*. The dashed line corresponds to the violation boundary. Any dot above the line is a violation since the actual error degradation is greater than the specified error tolerance.

Throughout the ten folds we evaluated we did not observe any accuracy tolerance violations with Tolerance Tiers, which reinforces confidence in the statistical guarantees provided by the routing rule generator. However, the `Naive-OSFA` policy results in several violations because it does not account for potential service request distribution variations. As a result, the selected service version deployed under `Naive-OSFA` is overfit to the training data. In contrast, the Tolerance Tier routing generator is able to create designs that generalize well to that variation. In many cases, we observe that the Tolerance Tiers have error degradations significantly less than what is afforded by the specified accuracy tolerance.

For the remainder of the section we do not consider `Naive-OSFA` since it results in violations. Instead, we use `TT-OSFA` as our baseline because it represents a one size fits all approach that provides the same guarantees as the ensembles that the routing generator can produce.

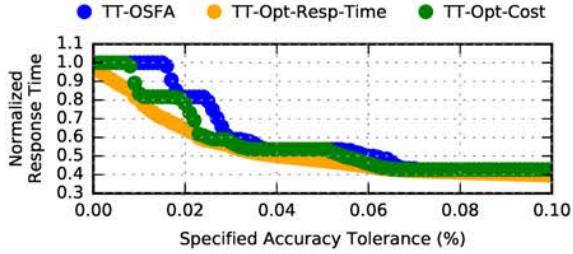
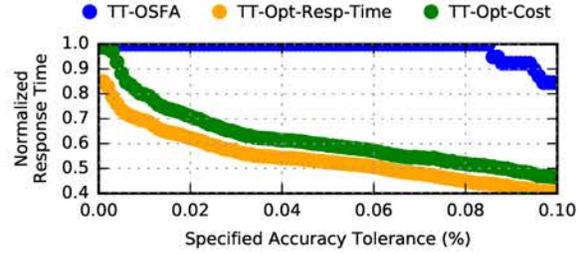

(a) ASR.  (b) Image Classification.

Figure 9: Response Time Results: The Tolerance Tier mechanisms can enable significant service response latency reductions compared to using a single service version for a given accuracy tolerance. It never performs worse.

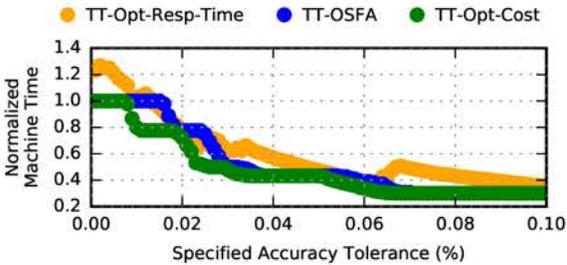
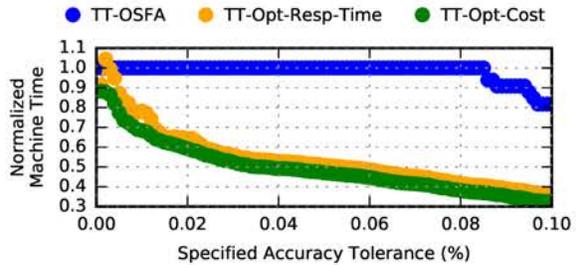

(a) ASR.  (b) Image Classification.

Figure 10: Infrastructure cost results. The Tolerance Tiers mechanisms can reduce machine processing time to lower infrastructure costs. The trade-off between the sequential and concurrent routing policies is between `TT-Opt-Cost` and `TT-Opt-Resp-Time`, respectively as the concurrent policy uses more resources to minimize time.

### D. Response Time

Fig. 9a and Fig. 9b show the response time improvements for the ASR and IC services. `TT-Opt-Resp-Time` is always able to outperform `TT-OSFA` because it utilizes the *concurrent* service version ensembling policy to produce a result. As discussed in Sec. IV-B, this is because the *concurrent* policy is able to hide the latency of utilizing multiple service versions, but as we will soon see comes at the expense of increased service invocation cost.

For a 1%, 5%, and 10% error tolerance, `TT-Opt-Resp-Time` outperforms the `TT-OSFA` by about 20% for the ASR service and the IC service can reach 20%, 40%, and even 50% reductions. This is because the most accurate IC service version is substantially more accurate than the next most accurate version. `TT-Opt-Cost` performs at least as good as `TT-OSFA` but is above to provide improvements at lower error tolerances than `TT-OSFA`. For example, `TT-Opt-Cost` provides the same response time at a 1% error degradation as `TT-OSFA` achieves at almost 2% in the ASR service. `TT-Opt-Cost` provides almost as much of a response time reduction as `TT-Opt-Resp-Time`.

### E. Machine Cost

Fig. 10a and Fig. 10b show the cost improvements for the ASR and IC services, respectively. `TT-Opt-Cost` performs at least as good as `TT-OSFA`, and typically provides lower invocation costs, for the ASR service. Due to the large difference in IC service version accuracies, `TT-Opt-Cost` significantly outperforms `TT-OSFA` as even tolerating *no accuracy degradation results in a 10% invocation cost reduction*.

At 1%, 5%, and 10% the invocation cost reductions for the IC service are roughly 35%, 55%, and 60%. This happens because early termination in the *sequential* ensemble routing poilicy is more heavily utilized for the IC service than the ASR service. The ASR service often requires full operation before selecting the best result, not so much that it doesn't provide a benefit in comparison to *OSFA*. `TT-Opt-Resp-Time` can exceed the invocation costs of `TT-OSFA` because of the excess computation that results from the service ensemble *concurrent* policy. However, this happens only at low tolerances for the IC tolerance. After a 0.5% error tolerance, `TT-Opt-Resp-Time` is able to achieve similar cost reductions as `TT-Opt-Cost` for the IC service.

## VI. Prior Work

**Cloud and Datacenter-scale Computing** Most cloud computing and datacenter-scale research aims to provide general purpose solutions to cloud computing problems from cluster management [28], [29], [30], cost-efficient resource provisioning [31], [32], workload interference and scheduling [33], [34], [35], heterogenous resource scheduling [36], [33], [34] and power management techniques [37]. While these works improve machine learning-based cloud workloads, they miss additional optimization opportunities domain-specific techniques can achieve. Tolerance Tiers demonstrates optimizations by exploiting the latency accuracy trade-off.

In addition, performance characteristics for datacenter workloads have been rigorously studied from a process (micro)architecture [38] and end-to-end cloud service [39] perspective. However, to the best of our knowledge, our work is the first to demonstrate the implications of the accuracy-latency relationship on the design of machine learning-based cloud services.

**Machine Learning Algorithms** Big/little neural networks [40] uses a fast network that can opt to use a more accurate network after. Our work generalizes this ensemble approach [41] to consider different policies and uses a general confidence metric that allows it to work with out machine learning applications beyond neural networks (e.g. ASR). At the algorithmic-level neural networks that terminate execution early have been proposed [42], [43]. MobileNets provide a parameterized training framework that can produce models with different latency-accuracy trade-offs. Another line of work compresses models, in memory size and computation complexity without sacrificing accuracy [44], [45].

**Model Serving Platforms** Platforms such as Velox [46], MCDNN [47], and Clipper [48] serve consumers with optimal models. However, some like the TensorFlow engine only serve one model at a time, or lack dynamic model selection, others such as [47], [48], [46] provide only one the policies described in this paper.

**Hardware Support for Machine Learning** Architecture research has put substantial focus on accelerating machine learning-based workloads, including neural networks [49], [50] and ASR [5]. These works are almost entirely focused on performance improvements and do not consider the accuracy-latency trade-off aspect of these workloads as we do. For example, [6] and [5] also optimize ASR beam searches but they do not evaluate their improvements on both latency and accuracy for different beam search parameters as we do.

**Approximate Computing** Approximation techniques can be static [51] or dynamic [52], [53]. Our work is the latter, treating training data as canary inputs [54]. At its core Tolerance Tiers is a domain specific approximate computing technique. Similar to our goals, MCDNN [47] proposes a system specific for approximation and applies neural networks on video streams where resources dictate the approximation level. In contrast, Tolerance Tiers allows for individual API consumers to dictate their accuracy requirements and it generalizes to many different machine learning applications.

## VII. Conclusion

As machine learning cloud services continue to be deployed at scale, it is important to investigate and deploy new cloud service architecture designs. Many of the machine learning techniques underlying today's machine learning-based cloud services exhibit similar accuracy-latency design trade-offs to the ASR and IC services we study, while still being deployed under the conventional "one size fits all" homogeneous deployment scheme. We show that optimizing accuracy without compromising responsiveness or cost is possible using *Tolerance Tiers*. Tolerance Tiers provide a means for intelligently exposing the inherent accuracy versus latency trade-off in machine learning cloud services to machine learning API consumers, shifting the power to the users by empowering MLaaS vendors with flexibility.

## VIII. Acknowledgements

We thank the reviewers for their useful feedback and candid opinions. This work was done in part while Matthew Halpern was a Ph.D. student at The University of Texas at Austin and a co-op at IBM Research. The work was funded by generous support from the National Science Foundation under CCF-1619283. Any opinions, findings, conclusions, or recommendations expressed in this article are those of the authors and do not necessarily reflect the views of their affiliations/institutions.